\documentclass{article} 
\usepackage{iclr2015,times}
\usepackage{hyperref}
\usepackage{url}
\usepackage{natbib}
\usepackage{amssymb,amsmath,amsfonts}
\usepackage{graphicx}

\title{Diverse Embedding Neural Network\\Language Models}

\author{
Kartik Audhkhasi$^*$, Abhinav Sethy$^*$ \& Bhuvana Ramabhadran \\
IBM T. J. Watson Research Center\\
Yorktown Heights, NY 10598, USA\\
\texttt{\{kaudhkha,asethy,bhuvana\}@us.ibm.com} \\
\footnotesize{\emph{$^*$Kartik Audhkhasi and Abhinav Sethy are joint first authors.}
}
}

\iclrfinalcopy 

\begin{document}

\maketitle

\begin{abstract}
We propose Diverse Embedding Neural Network (DENN), a novel architecture for language models (LMs). A DENNLM projects the input word history vector onto multiple diverse low-dimensional sub-spaces instead of a single higher-dimensional sub-space as in conventional feed-forward neural network LMs. We encourage these sub-spaces to be diverse during network training through an augmented loss function. Our language modeling experiments on the Penn Treebank data set show the performance benefit of using a DENNLM.
\end{abstract}

\section{Introduction}\label{sec:intro}
 Diversity of systems trained to perform a given machine learning task is
crucial to obtaining a performance improvement upon fusion (\cite{kuncheva2004combining}). The problem of language modeling, that aims to design predictive models of text, 
 is no exception. Several language models (LMs) have been proposed over many years of research (\cite{rosenfield2000two}). 
 Simple N-gram models estimate the conditional probability of the $i$-th word $w_i$ in a sentence given the previous $N-1$ words $(w_{i-N+1},\ldots,w_{i-1})$ as
\begin{align}
	\hat{P}_\text{N-gram}(w_i|w_{i-N+1},\ldots,w_{i-1}) &= \frac{C(w_i,w_{i-1},\ldots,w_{i-N+1})}{C(w_{i-1},\ldots,w_{i-N+1})}
\end{align} 
where $C(.)$ computes the count of a given word phrase or N-gram in the training text\footnote{Almost all N-gram models are smoothed in various ways to assign non-zero probability estimates for word phrases unseen in training data. We use Kneser-Ney smoothing (\cite{kneser1995improved}) for all N-grams models in this paper.}. More complex LMs such as feed-forward neural networks (NNs) (\cite{bengio2006neural}) estimate this probability as a non-linear function
\begin{align}
	\hat{P}_\text{NN}(w_i|w_{i-N+1},\ldots,w_{i-1}) &= f(w_{i-1},\ldots,w_{i-N+1};\Theta)
\end{align}
parameterized by $\Theta$\footnote{We will describe the typical architecture of
a NNLM in the next section}. Researchers have found that fusing different kind
of  n-gram language models
together (\cite{Goodman01abit},~\cite{mikolov2011empirical}) often significantly
improves performance.
Table~\ref{tab:upenn_baselines} shows the perplexity\footnote{Perplexity is a
standard measure of LM performance and is $2^{-L}$ where $L$ is the negative
average log-likelihood of the test set text estimated by the LM.} of 4-gram and
NNLMs on a standard split of the Penn Treebank data set
(\cite{marcus1993building}). Interpolation of a NNLM with a 4-gram LM gives a
16.9\% reduction in perplexity over a single NNLM even though the two LMs have
relatively close perplexities. We use the correlation coefficient between the posterior
probabilities of the predicted word over the test set from the two models
as a simple measure to predict whether the models are diverse. If the
posterior probabilities are highly correlated, then the models are less diverse and
smaller gains are expected from fusion. The posteriors from the N-gram and NNLM
have a correlation coefficient of 0.869 which is significantly lower than the correlation coefficient of 0.988 for a pair of
 randomly initialized NNLMs. This higher diversity of the NNLM and N-gram LM
 combination results in
 significant perplexity improvement upon interpolation.

\begin{table}[h]
\label{tab:upenn_baselines}
\caption{This table shows the test set perplexities of several LMs on the Penn Treebank test set. A (X,Y) N-gram NNLM projects the one-hot vector\protect\footnotemark of the previous N-1 words onto an X-dimensional linear sub-space. 
It then inputs this projected vector into a NN with one hidden layer with Y neurons that outputs the posterior probability of 
all words in the vocabulary. $K$ NN-RandInit denotes interpolation of $K$ randomly initialized and independently-trained NNLMs. 
The last column shows the average correlation coefficient between posteriors of the test set words estimated by the LMs in the ensemble.}
	\begin{center}
	\begin{tabular}{|c|c|c|c|}
	\hline
	{\bf LM Name} & {\bf Description} & {\bf Perplexity} & {\bf Posterior} \\
	 & & & {\bf Corr. Coeff.} \\ \hline\hline
	4-gram & 4-gram Kneser-Ney smoothed & 142.04 & N/A \\ \hline
	1 NN & 1x(600,800) 4gm NN-LM & 140.06 (-1.4\%) & N/A \\ \hline
	4 NN-RandInit & Randomly initialized & 134.80 (-5.1\%) & 0.99 \\
	 & 4x(150,200) 4gm NN-LMs & & \\ \hline
	4-gram + 1 NN & Interpolation & 116.33 (-18.1\%) & 0.87 \\ \hline
	4-gram + 4 NN-RandInit & Interpolation & 116.17 (-18.2\%) & 0.88 \\ \hline
	\end{tabular}	
	\end{center}
\end{table}
\footnotetext{A one-hot vector of the $i$-th word in the vocabulary contains $1$ at index $i$ and $0$ everywhere else.}

Random initialization and modifying the neural net topology in terms of the
embedding and hidden layer size can be used to build diverse NNLM models. As we
can see from Table~\ref{tab:upenn_baselines}, 4 randomly initialized NNLMs (4
NN-RandInit) when fused together provide a 5\% improvement in perplexity over
the baseline. Recurrent NNLM models of different topologies can be fused to get
significant gains as well as demonstrated in~(\cite{mikolov2011empirical}). The
remarkable benefit of such simple diversity-promoting strategies leads us to the central question of this paper - Is there a way to explicitly enforce diversity during NNLM model training? We show that modifying the NNLM architecture and augmenting the training loss function achieves that. The fact that a NNLM learns a low-dimensional continuous space embedding of input words motivates the architecture and training of our proposed model - Diverse Embedding Neural Network (DENN) LM.

We first give an overview of conventional NNLMs in the next section. Section~\ref{sec:dennlm} presents the DENNLM architecture and its training loss function. We presents experiments and results in Section~\ref{sec:expts} and conclude the paper in Section~\ref{sec:concl}.

\section{Feed-forward Neural Network Language Model (NNLM)}\label{sec:nnlm}
\begin{figure}[h]
	\begin{center}
		\includegraphics[scale=0.25]{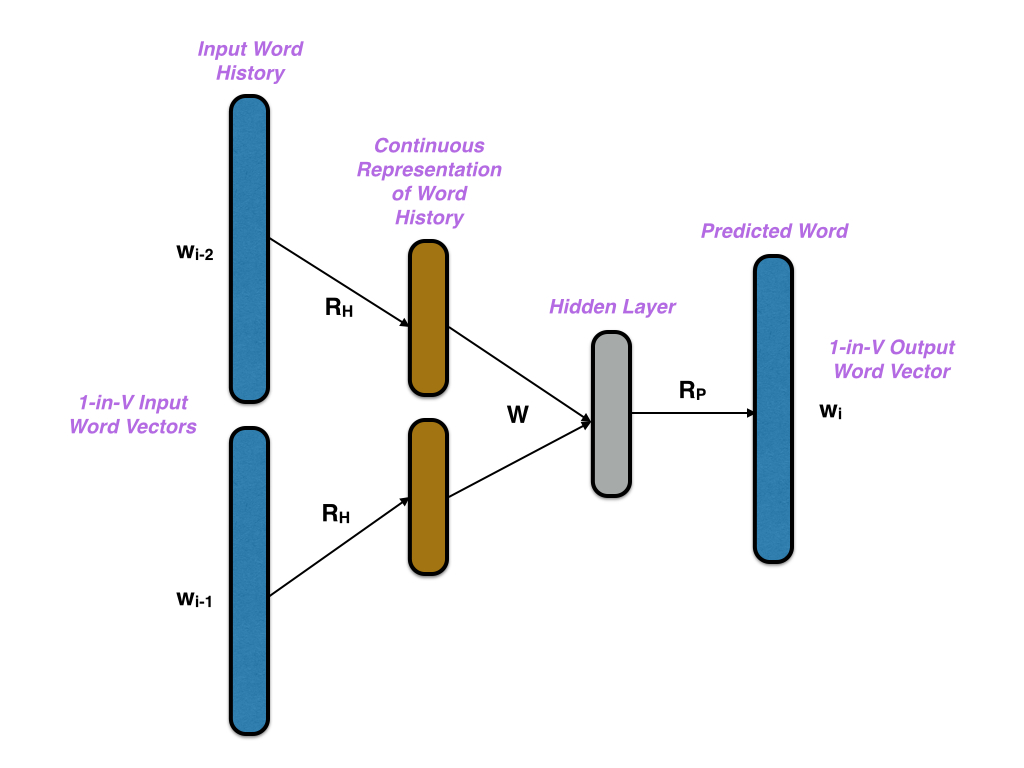}
	\end{center}
	\caption{This figure shows the schematic diagram of a 3-gram NNLM. The matrix $\mathbf{R}_H$ projects the input history word vectors onto a continuous space. These representations then pass through a NN with one hidden layer to predict the next word.}
	\label{fig:1NN}
\end{figure}

A feed-forward neural network LM (NNLM) converts the one-hot encoding of each word in the history to
a continuous low-dimensional vector representation (\cite{bengio2006neural}). Figure~\ref{fig:1NN} shows the schematic diagram
of a typical NNLM. Let $\mathbf{w}_i,\ldots,\mathbf{w}_{i-N+1}$ denote
the $1$-in-$V$ vectors of the history words. Let $\mathbf{R}_H$ denote the $D \times V$ matrix that projects a history word vectors onto $D$-dimensional vectors $\mathbf{R}_H\mathbf{w}_i,\ldots,\mathbf{R}_H\mathbf{w}_{i-N+1}$. $D$ is typically much smaller than the size $V$ of the vocabulary with typical
values being $V = 10,000$ and $D = 500$.

This resulting $(N-1)D$-dimensional continuous representation
of input words feeds into a neural network with one hidden layer that estimates the $1$-in-$V$ vector
of the target word $w_i$. The hidden neuron activation function is hyperbolic tangent or logistic sigmoid while the output neuron activation function is a $V$-way softmax. The back-propagation algorithm (\cite{rumelhart1986learning}) trains an NNLM, often using stochastic gradient descent where the gradient is computed over several random subsets or batches of N-grams from the input training data.

Researchers have proposed several variants of the feed forward NNLM, especially
to model a longer input word history. Prominent examples include the recurrent neural
 network LM (\cite{mikolov2010recurrent}) and bidirectional long-short term memory (LSTM) LM (\cite{sundermeyer2012lstm}). 
 These models offer improved performance but can be slower and more difficult to
 train. Even though we restrict our attention to feed forward NNLMs in this
 paper, the general principles proposed are applicable to other NNLM
 architectures as well.
 The next section introduces the diverse embedding NNLM (DENNLM).

\section{Diverse Embedding NNLM}\label{sec:dennlm}
A diverse embedding NNLM (DENNLM) aims to learn multiple diverse representations of the input words rather than
a single representation. It is first important to understand the intuition of a representation in the context
of a NNLM. Given a set of $N$ input word vectors $\mathbf{w}_{i-1},\ldots,\mathbf{w}_{i-N+1}$, consider
the set of $D$-dimensional vectors $\mathbf{R}_H\mathbf{w}_{i-1},\ldots,\mathbf{R}_H\mathbf{w}_{i-N+1}$.
The pairwise distances between vectors of this set constitute the representation of the input words. A good representation captures the contextual and semantic similarity between pairs of words. Similar words are located close to each other in representation while dissimilar words are located far apart. 
A NNLM uses this representation of the input words to predict the next word. The most natural way to ensure diversity of two NNLMs is through the diversity in the representation itself (representational diversity). The next section
discusses an intuitive score to capture representational diversity.

\subsection{Diversity between NNLM Embeddings}\label{sec:div_types}
Maximizing the representational diversity between two NNLMs first requires an objective score to capture this diversity. Consider the representation $\mathbf{R}_H\mathbf{w}_{i-1},\ldots,\mathbf{R}_H\mathbf{w}_{i-N+1}$ of the $N-1$ input words to the NNLM. Since this representation lies in a $D$-dimensional Euclidean space, the set of all $(N-1)(N-2)/2$ pairwise angles
 between $N-1$ words are sufficient to uniquely define the representation even under any affine transformation such as translation, rotation, and scaling. The matrix of pairwise cosine angles between all pairs of points in the representation defined by $\mathbf{R}_{H1}$ is

\[ \mathbf{C}(\mathbf{R}_{H1}) = \left( \begin{array}{ccc}
1 & \ldots & \frac{\mathbf{w}_{i-1}^T\mathbf{R}_{H1}^T\mathbf{R}_{H1}\mathbf{w}_{i-N+1}}{||\mathbf{R}_{H1}\mathbf{w}^{i-1}||~||\mathbf{R}_{H1}\mathbf{w}^{i-N+1}||} \\
\frac{\mathbf{w}_{i-2}^T\mathbf{R}_{H1}^T\mathbf{R}_{H1}\mathbf{w}_{i-1}}{||\mathbf{R}_{H1}\mathbf{w}^{i-2}||~||\mathbf{R}_{H1}\mathbf{w}^{i-2}||} & \ldots & \frac{\mathbf{w}_{i-2}^T\mathbf{R}_{H1}^T\mathbf{R}_{H1}\mathbf{w}_{i-N+1}}{||\mathbf{R}_{H1}\mathbf{w}^{i-2}||~||\mathbf{R}_{H1}\mathbf{w}^{i-N+1}||} \\
\vdots & \ddots & \vdots \\
\frac{\mathbf{w}_{i-N+1}^T\mathbf{R}_{H1}^T\mathbf{R}_{H1}\mathbf{w}_{i-1}}{||\mathbf{R}_{H1}\mathbf{w}^{i-N+1}||~||\mathbf{R}_{H1}\mathbf{w}^{i-1}||} & \ldots & 1
\end{array} \right).\]

This matrix completely defines the representation of the word history produced by $\mathbf{R}_{H}$.
It is also independent of the dimensionality of the representation, i.e. the number of rows of $\mathbf{R}_H$. This is useful when comparing two representations with different dimensionality of the same set of input data points. Given two such representations computed using matrices $\mathbf{R}_{H1}$ and $\mathbf{R}_{H2}$, we define the representational diversity as the negative correlation between cosine angles
\begin{align}
	d_\text{Rep}(\mathbf{C}(\mathbf{R}_{H1}),\mathbf{C}(\mathbf{R}_{H2})) &= -vec\Big{(}\mathbf{C}(\mathbf{R}_{H1})\Big{)}^Tvec\Big{(}\mathbf{C}(\mathbf{R}_{H2})\Big{)}
\end{align}
across the two representations, where $vec$ raster-scans a matrix into a column vector. We note that $d_\text{Rep}$ is bounded because the cosine is bounded between $-1$ and $1$. 
Representational diversity between $\mathbf{R}_{H1}$ and $\mathbf{R}_{H2}$
increases as $d_\text{Rep}$ decreases. In our implementation of distance
diversity, for computation efficiency reasons we consider distances over a
randomly chosen set of 500 words in each minibatch instead of the full vocabulary. Our experiments show that using the entire vocabulary for diversity computation gave only minor improvements in perplexity at the expense of much longer training time.

We are currently exploring several other potential ways to compute diversity between two NNLMs beyond the representational diversity score presented in this paper. 
The next section discusses the DENNLM architecture and training loss function.


\subsection{DENNLM Architecture and Training Loss Function}\label{sec:dennlm_training}
\begin{figure}[h]
	\begin{center}
		\includegraphics[scale=0.25]{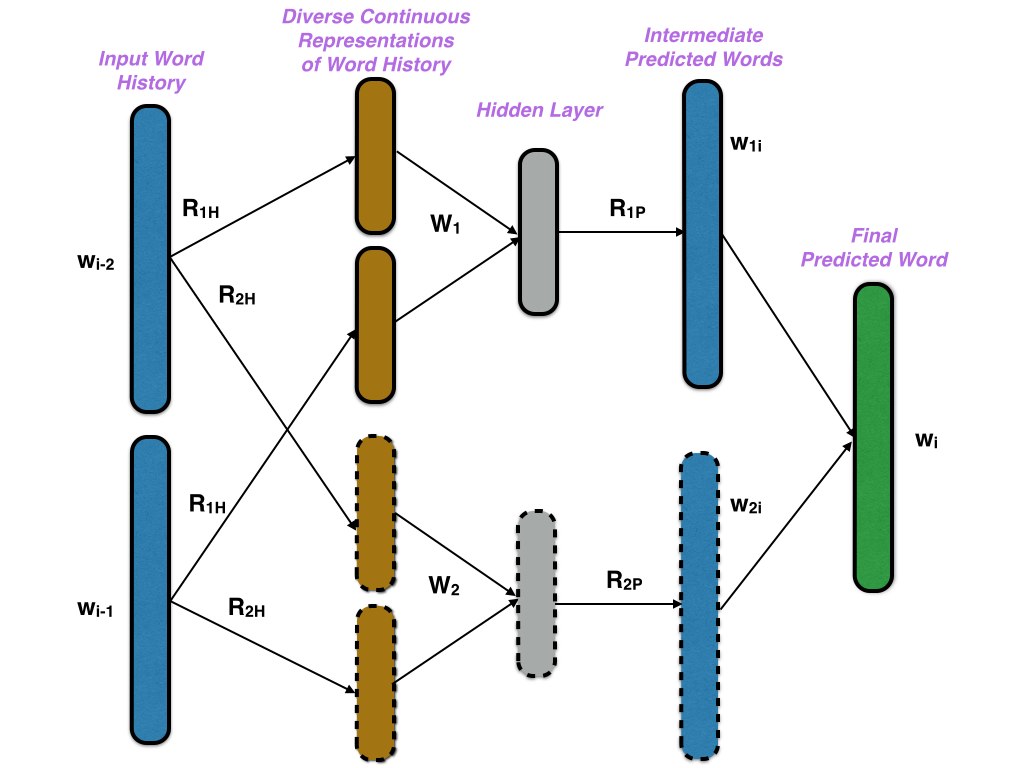}
	\end{center}
	\caption{This figure shows the schematic diagram of a 3-gram DENNLM with two diverse representations.}
	\label{fig:DENNLM}
\end{figure}

As discussed earlier, a DENNLM attempts to learn multiple diverse low-dimensional representations of the
input word history. Figure~\ref{fig:DENNLM} shows the schematic diagram of a DENNLM with two diverse representations. The two
representations pass through two separate NNs and produce separate predictions of the next word. The model
merges the two predictions to produce the final prediction.

It is important to note that
the DENNLM is equivalent to a single NNLM with the following block-structured representation matrix

\[ \mathbf{R}_{H} = \left( \begin{array}{cc}
\mathbf{R}_{1H} & \mathbf{0} \\
\mathbf{0} & \mathbf{R}_{1H} \\
\mathbf{R}_{2H} & \mathbf{0} \\
\mathbf{0} & \mathbf{R}_{2H}
\end{array} \right)\]
and block-diagonal connection weight matrices
\[ \mathbf{W} = \left( \begin{array}{cc}
\mathbf{W}_{1} & \mathbf{0} \\
\mathbf{0} & \mathbf{W}_{2}
\end{array} \right) \quad
\mathbf{R}_{P} = \left( \begin{array}{cc}
\mathbf{R}_{1P} & \mathbf{0} \\
\mathbf{0} & \mathbf{R}_{2P}
\end{array} \right) \;.
\]

The presence of zero entries in the above matrices implies that a DENNLM has a smaller number of parameters than a comparable NNLM. The equivalence between a DENNLM and a conventional NNLM makes the implementation of a DENNLM easy within
conventional NNLM toolkits.

Merely constraining the architecture of a NNLM does not guarantee that it learns multiple diverse representations. Hence we augment the training loss function of a conventional NNLM with additional terms to promote diversity. We use the negative log-likelihood or cross entropy for $V$-way classification as the loss function
for a conventional NNLM. A DENNLM instead uses the following augmented loss function:
\begin{align}\label{eq:dennlm_loss}
	L_\text{DENNLM} &= -\beta\sum_{i=1}^T \log \Big{(}\sum_{m=1}^M \alpha_m p_m(w_i|w_{i-1},\ldots,w_{i-N+1})\Big{)} \nonumber \\ 
	&- (1-\beta)\sum_{i=1}^T \sum_{m=1}^M \alpha_m \log \Big{(} p_m(w_i|w_{i-1},\ldots,w_{i-N+1})\Big{)} \nonumber \\
	&- \gamma d_\text{Rep}(\mathbf{C}(\mathbf{R}_{H1}),\ldots,\mathbf{C}(\mathbf{R}_{HM})) 
\end{align}
The first term of the
above loss function is a mixture model loss that ensures that the fused prediction is
accurate.  Using only the first term trains a mixture of neural networks.
However, this will not ensure high discrimination ability of the representations learned by the individual networks 
because the different representations will only capture the modes of the data
distribution. We thus include the second term, motivated by a recent work on deeply supervised neural
networks~(\cite{deeplysupervised}), where the authors 
augment the conventional loss at the final layer with discriminative loses computed at
previous layers.
The second term in (\ref{eq:dennlm_loss}) plays a similar role and makes the individual representations 
discriminative as well.  The final term
is the representational diversity of the NNLM as described in
Section~\ref{sec:div_types}.
Minimizing the DENNLM loss function in (\ref{eq:dennlm_loss}) gives representations that are jointly discriminative, individually discriminative, and diverse.

We can add $L_1$ and/or $L_2$ regularization penalties to the loss function as well, which are often useful to prevent over-fitting in case the network size is large compared to the
number of training set N-grams. The next section discusses the experimental setup and results.

\section{Experiments and Results}\label{sec:expts}
We conducted language modeling experiments on a standard split of the Penn Treebank (UPenn) data set (\cite{marcus1993building}), as used in previous works such as (\cite{mikolov2011empirical}). The UPenn data set has a vocabulary of 10K words, and contains approximately 1M N-grams in the training set. We used the UPenn data set since it is well-studied for language modeling and also small enough to conduct several experiments for understanding the DENNLM model better. We
implemented the DENNLM in Theano (\cite{theano}) and trained it by
optimizing the augmented loss function in (\ref{eq:dennlm_loss}) using Root-Mean Square Propagation (RMSProp)~(\cite{rmsprop}), a variant of stochastic gradient descent. We tuned all hyper-parameters on the standard UPenn held-out set.

Table~\ref{tab:dennlm_results} shows the test set perplexities of the baseline and the DENNLMs. We kept
the NN model size comparable by reducing the size of each component NNLM. Our results show a significant
improvement in perplexity by using a DENNLM over the 4-gram model, a single NNLM, and interpolation of randomly initialized NNLMs.
 The posterior correlation coefficients are significantly less than 0.99, which is the correlation coefficient for randomly 
 initialized NNLMs.  

\begin{table}[h]
\label{tab:dennlm_results}
\caption{This table shows the test set perplexities of baseline and DENNLMs on the Penn Treebank test set.}
	\begin{center}
	\begin{tabular}{|c|c|c|c|}
	\hline
	{\bf LM Name} & {\bf Description} & {\bf Perplexity} & {\bf Posterior} \\
	 & & & {\bf Corr. Coeff.} \\ \hline\hline
	4-gram & 4-gram Kneser-Ney smoothed & 142.04 & N/A \\ \hline
	1 NN & 1x(600,800) 4gm NN-LM & 137.32 (-3.3\%) & N/A \\ \hline
	4 NN-RandInit & Randomly initialized & 134.85 (-5.1\%) & 0.99 \\
	 & 4x(150,200) 4gm NN-LMs & & \\ \hline
	4 DENN & Diversely trained & 122.69 (-13.6\%) & 0.89  \\
	 & 4x(150,200) 4gm NN-LMs &  & \\ \hline
	8 DENN & Diversely trained & 116.41 (-18.0\%) & 0.83  \\
	 & 8x(75,100) 4gm NN-LMs & & \\ \hline
	4-gram + 1 NN & Interpolation & 116.33 (-18.1\%) & 0.87 \\ \hline
	4-gram + 4 DENN & Interpolation & 112.79 (-20.6\%) & 0.88 \\ \hline
	4-gram + 8 DENN & Interpolation & 109.32 (-23.0\%) & 0.92 \\ \hline
	\end{tabular}	
	\end{center}
\end{table}

Note that  the DENNLM significantly outperforms a standard NNLM of size
(600,800) which has similar number of parameters. It is also clearly better then
a randomly initialized set of 4 NNLM models. The perplexity results in
Table~\ref{tab:dennlm_results} for our diverse feed-forward NNLMs are especially encouraging given the fact that the more advanced recurrent neural network (RNN) LM gives a perplexity of 124.7 by 
itself and 105.7 upon interpolation with a 5-gram LM on the same task (\cite{mikolov2011empirical}).


\subsection{Sensitivity to Hyperparameters}\label{sec:hyper}
\begin{figure}[h]
	\begin{center}
		\includegraphics[scale=0.45]{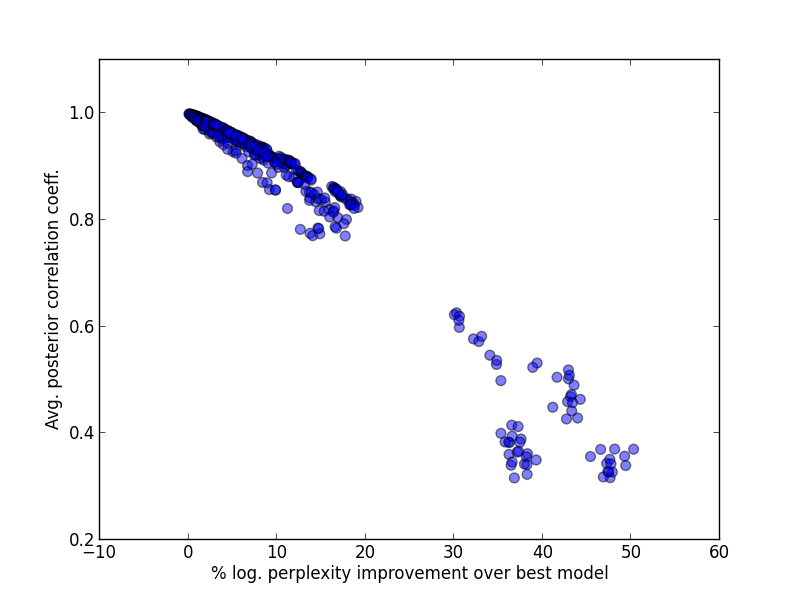}
	\end{center}
	\caption{This figure shows the scatter plot between average posterior cross-correlation and percent improvement of the DENNLM log perplexity over the best model's log perplexity. We observed a strong correlation coefficient of -0.97 between the two variables indicating that more diverse NNLMs give a bigger reduction in perplexity upon interpolation.}
	\label{fig:pcc_ppl_scatter}
\end{figure}

We further studied the dependence of DENNLM performance on some sets of hyper-parameters and list our observations below:
\begin{itemize}
  \item \emph{Diversity parameters}: The performance was fairly stable over a range of
  $\beta$ between 0.3 to 0.7 with minor differences of around 2 points in perplexity.
  The performance with respect to diversity weight $\gamma$ was also stable in the range 1 to 3
  \item \emph{Model parameters}: We explored different model sizes for both the single
  model and DENNLM by changing the DENNLM topology (number of hidden neurons, layers etc.). We observed that in general, a DENNLM with the same total size
  as a single NNLM works well. We can achieve smaller gains of around 3 to 4 points in perplexity
  by increasing the size of the DENNLM.
\item \emph{Optimization parameters}: We observed that the choice of the optimization algorithm can have
a significant impact on the diversity of models generated with random initialization. We found that using
 RMSProp with a significantly higher learning rate, gradient clipping, and
 scaling of bias updates while training four randomly-initialized NNLMs
 leads to a lower posterior posterior correlation of 0.95 compared to 0.99 with our
 standard optimization setting. This higher diversity between models translates to a lower
 perplexity of 120 with the interpolated NNLM which is comparable to our best results in
 Table~\ref{tab:dennlm_results}. This indicates that the choice of hyper-parameters and optimizer settings for
 building diverse models via random initialization can be different from the
 ones used for training the models individually.
 \end{itemize}
To further understand the impact of hyper-parameters on DENNLM diversity and perplexity, we performed
a thorough grid search of $\beta$ and $\gamma$ in (\ref{eq:dennlm_loss}), the learning rate of RMSProp, and the weight of an L2 penalty on the DENNLM connection weights. We then computed the log perplexity
and average cross-correlation between posteriors of individual models in a DENNLM. Figure~\ref{fig:pcc_ppl_scatter} shows the scatter plot between average posterior cross-correlation and percent improvement of the DENNLM log perplexity over the best model's perplexity. A strong negative correlation of $-0.97$ in this figure shows that more diverse models give a bigger improvement in log perplexity upon interpolation. This highlights the merit of training diverse NNLM and the fact that
one can achieve this diversity by either informed objective functions such as the one in (\ref{eq:dennlm_loss}) or an exhaustive hyper-parameter search. The latter becomes especially tedious for deep and complex neural networks trained on big data sets.

\section{Conclusion}\label{sec:concl}
In this work, we introduced a neural network architecture and training objective function that encourages the individual models to be diverse in terms of their output distribution as well as the underlying word representations. We demonstrated its usefulness on the well-studied UPenn language modeling task. The proposed training criterion is general enough and does not constrain the NNLM models to be of any specific architecture. Given the promising results, our next step is to evaluate the improved language models on speech recognition and spoken-term detection tasks. We also plan to explore the utility of these diverse representations to measure semantic similarity and for sentence completion, where word embedding-based models have been shown to be effective.

\section{Acknowledgement}\label{sec:ack}
This work was supported by the Intelligence Advanced Research Projects Activity (IARPA) via Department of Defense U.S.
Army Research Laboratory (DoD / ARL) contract number W911NF-12-C-0012.  The U.S. Government is
authorized to reproduce and distribute reprints for Governmental purposes notwithstanding any copyright
annotation thereon. Disclaimer: The views and conclusions contained herein are those of the authors and should not
be interpreted as necessarily representing the official policies or endorsements, either expressed or implied, of
IARPA, DoD/ARL, or the U.S. Government.


\bibliography{diverse_NNLM}
\bibliographystyle{iclr2015}

\end{document}